\documentclass[10pt,twocolumn,letterpaper]{article}

\newif\ifarxiv
\arxivtrue

\ifarxiv
    \usepackage[pagenumbers]{cvpr}              
\else
    \usepackage[review]{cvpr}      
\fi

\usepackage{enumitem}
\usepackage{graphicx}
\usepackage{amsmath}
\usepackage{amssymb}
\usepackage{booktabs}
\usepackage[accsupp]{axessibility}
\usepackage[pagebackref,breaklinks,colorlinks]{hyperref}
\usepackage{xcolor}
\usepackage[capitalize]{cleveref}
\crefname{section}{Sec.}{Secs.}
\Crefname{section}{Section}{Sections}
\Crefname{table}{Table}{Tables}
\crefname{table}{Tab.}{Tabs.}
\newcommand{\ett}{EgoT2}

\newcommand{\revision}[1]{\textcolor{black}{#1}}
\usepackage{makecell}
\usepackage{multirow}
\usepackage{arydshln}
\usepackage{dblfloatfix}
\def\cvprPaperID{6059} 
\def\confName{CVPR}
\def\confYear{2023}

\begin{document}

\title{Egocentric Video Task Translation}


\author{%
Zihui Xue\textsuperscript{1,2}\thanks{Work done during an internship at FAIR, Meta AI.} \quad Yale Song\textsuperscript{2} \quad Kristen Grauman\textsuperscript{1,2} \quad Lorenzo Torresani\textsuperscript{2} \\
\textsuperscript{1}The University of Texas at Austin \qquad
\textsuperscript{2}FAIR, Meta AI\\
}

\maketitle
\begin{abstract}
Different video understanding tasks are typically treated in isolation, and even with distinct types of curated data (e.g., classifying sports in one dataset, tracking animals in another).
However, in wearable cameras, the immersive egocentric perspective of a person engaging with the world around them presents an interconnected web of video understanding tasks---hand-object manipulations, navigation in the space, or human-human interactions---that unfold continuously, driven by the person's goals.  
We argue that this calls for a much more unified approach.  We propose EgoTask Translation (EgoT2), which takes a collection of models optimized on separate tasks and learns to translate their outputs for improved performance on any or all of them at once.  
Unlike traditional transfer or multi-task learning, EgoT2's ``flipped design'' entails separate task-specific backbones and a task translator shared across all tasks, which captures synergies between even heterogeneous tasks and mitigates task competition.   
Demonstrating our model on a wide array of video tasks from Ego4D, we show its advantages over existing transfer paradigms and achieve top-ranked results on four of the Ego4D 2022 benchmark challenges.\footnote{Project webpage: \url{https://vision.cs.utexas.edu/projects/egot2/}.}  

\end{abstract}



\section{Introduction}

In recent years, the introduction of large-scale video datasets (\eg, Kinetics~\cite{kay2017kinetics,carreira2017quo} and Something-Something~\cite{goyal2017something}) have enabled the application of powerful deep learning models to video understanding and have led to dramatic advances. These third-person datasets, however, have overwhelmingly focused on the single task of action recognition in trimmed clips~\cite{diba2020large,zhu2020comprehensive,ozyer2021human,kong2022human}. Unlike curated third-person videos, our daily life involves frequent and heterogeneous interactions with other humans, objects, and environments in the wild. First-person videos from wearable cameras capture the observer's perspective and attention as a continuous stream.  As such, they are better equipped to reveal these multi-faceted, spontaneous interactions. Indeed egocentric datasets, such as EPIC-Kitchens~\cite{damen2020epic} and Ego4D~\cite{ego4d}, provide suites of tasks associated with  varied interactions. However, while these benchmarks have promoted a broader and more heterogeneous view of video understanding, they risk perpetuating the fragmented development of models specialized for each individual task. 

\begin{figure}[!t]
  \centering
  \includegraphics[width=1.0\linewidth]{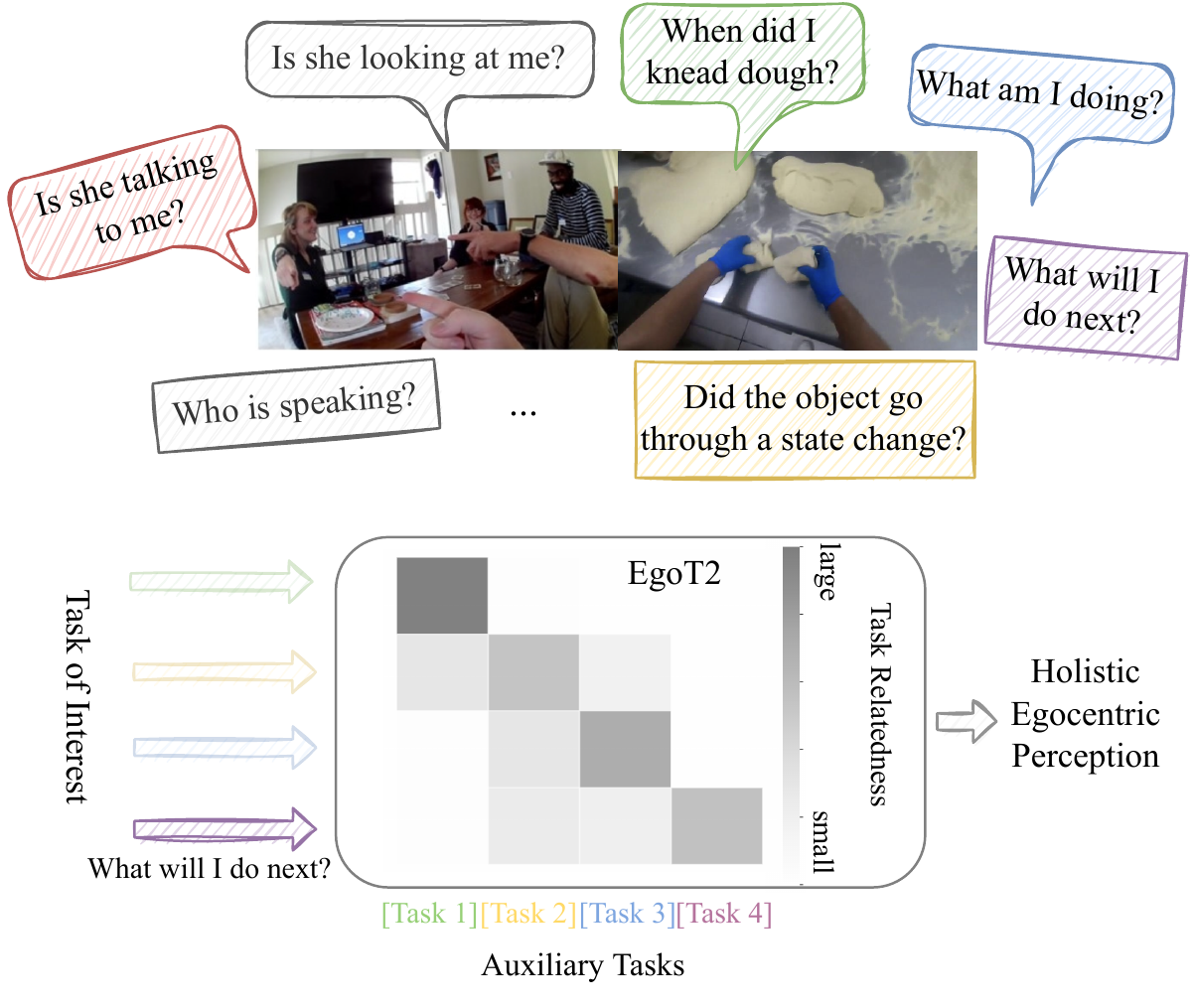}
  \caption{Given a set of diverse egocentric video tasks, the proposed \ett \ leverages synergies among the tasks to improve each individual task performance. The attention maps produced by \ett \ offer good interpretability on inherent task relations.}
  \label{fig:intro}
\end{figure}

In this work, we argue that the egocentric perspective offers an opportunity for \emph{holistic perception} that can beneficially leverage synergies among video tasks to solve all problems in a unified manner. See Figure~\ref{fig:intro}.

Imagine a cooking scenario where the camera wearer actively interacts with objects and other people in an environment while preparing dinner. These interactions relate to each other: a hand grasping a knife suggests the upcoming action of cutting; the view of a tomato on a cutting board suggests that the object is likely to undergo a state transition from whole to chopped; the conversation may further reveal the camera wearer's ongoing and planned actions. Apart from the natural relation among these tasks, 
egocentric video's \emph{partial observability} (\ie, the camera wearer is largely out of the field of view) further motivates us to seek synergistic, comprehensive video understanding 
to leverage complementary cues among multiple tasks.

Our goal presents several technical challenges for 
conventional transfer learning (TL)~\cite{zhuang2020comprehensive} and multi-task learning (MTL)~\cite{mtlsurvey}. 
First, 
MTL requires training sets where each sample includes annotations for all tasks~\cite{zamir2018taskonomy,pal2019zero,dwivedi2019representation,standley2020tasks,guo2020learning,sun2020adashare}, which is often impractical. 
Second, 
egocentric video tasks are heterogeneous in nature,  
requiring different modalities (audio, visual, motion), diverse labels (\eg, temporal, spatial or semantic), and different temporal granularities (\eg, action anticipation requires long-term observations, but object state recognition operates at a few sparsely sampled frames)---all of which makes a unified model design problematic and fosters specialization. 
Finally, while existing work advocates the use of a shared encoder across tasks to learn general representations~\cite{mtloverview,ma2016going,luvizon2020multi,fathi2012learning,li2018eye,huang2020mutual,baradel2018object,kapidis2019multitask}, the diverse span of egocentric tasks poses a hazard to parameter sharing which can lead to negative transfer~\cite{gong2019comparison,leang2020dynamic,standley2020tasks,guo2020learning}.

To address the above limitations, we propose EgoTask Translation (\ett), a unified learning framework to address a diverse set of egocentric video tasks together. 
\ett \ is flexible and general in that it can handle separate datasets for the different tasks; it takes video heterogeneity into account; and \revision{it mitigates negative transfer when tasks are not strongly related.}
To be specific, \ett \ consists of specialized models developed for individual tasks and a \emph{task translator} that explicitly models inter-task and inter-frame relations. 
We propose two distinct designs: (1) task-specific \ett \ (\ett-s) optimizes a given primary task with the assistance of auxiliary tasks (Figure~\ref{fig:related}(c)) while (2) task-general \ett \ (\ett-g) supports task translation for multiple tasks at the same time (Figure~\ref{fig:related}(d)).

Compared with a unified backbone across tasks~\cite{zamir2018taskonomy}, adopting task-specific backbones preserves peculiarities of each task (\eg different temporal granularities) and mitigates negative transfer since each backbone is optimized on one task. 
Furthermore, unlike traditional parameter sharing
~\cite{mtloverview}, the proposed task translator learns to ``translate'' all task features into predictions for the target task by selectively activating useful features and discarding irrelevant ones. The task translator also facilitates 
interpretability by explicitly revealing which temporal segments and which subsets of tasks contribute to improving a given task. 

We evaluate \ett~on a diverse set of 7 egocentric perception tasks from the world's largest egocentric video benchmark, Ego4D~\cite{ego4d}. Its heterogeneous tasks 
extend beyond mere action recognition to  
speaker/listener identification, keyframe localization, object state change classification, long-term action anticipation, and others, 
and provide a perfect fit for our study. 
Our results reveal inherent task synergies, demonstrate consistent performance improvement across tasks, and offer good interpretability in task translation. Among all four Ego4D challenges involved in our task setup, \ett \ outperforms all submissions to three Ego4D-CVPR'22 challenges and achieves state-of-the-art performance in one Ego4D-ECCV'22 challenge.

\section{Related Work}
\begin{figure}[!t]
  \centering
  \includegraphics[width=1.0\linewidth]{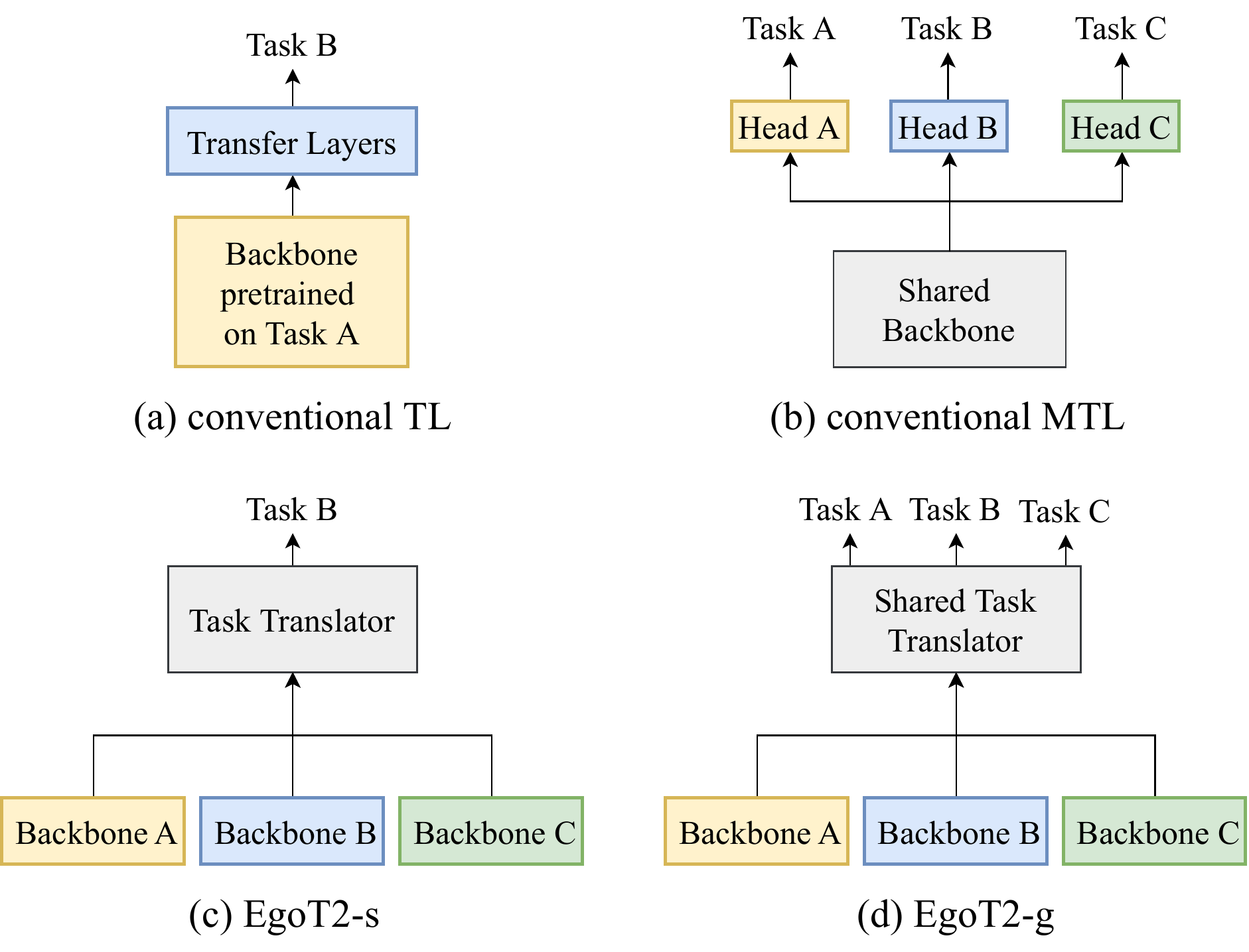}
  \vspace*{-0.2in}
  \caption{(a) Conventional TL uses a backbone pretrained on the source task followed by a head transferring supervision to the target task; (b) Traditional MTL consists of a shared backbone and several task-specific heads; (c) \ett-s adopts task-specific backbones and optimizes the task translator for a given primary task; (d) \ett-g jointly optimizes the task translator for all tasks.}
  \label{fig:related}
\end{figure}

\textbf{Transfer Learning}. TL~\cite{zhuang2020comprehensive} aims at transferring knowledge from a source domain to improve the performance in a target domain. The most widely adopted approach is to pretrain a model on a source task then finetune on the target task, as shown in Figure \ref{fig:related}(a). Following this paradigm, many video classification models~\cite{liu2022video,bertasius2021space,wang2016temporal,arnab2021vivit} are initialized from models pretrained on ImageNet~\cite{deng2009imagenet}. In addition, many works propose to transfer knowledge from a large-scale video dataset (\eg, Kinetics~\cite{kay2017kinetics,carreira2017quo}) to benefit action recognition in smaller-scale datasets~\cite{stroud2020d3d} such as UCF-101~\cite{soomro2012ucf101} and HMDB-51~\cite{kuehne2011hmdb} or to improve other video tasks, such as spatiotemporal action localization~\cite{iqbal2019enhancing,fan2021multiscale,feichtenhofer2019slowfast,pan2021actor, arnab2022beyond} and video anomaly detection~\cite{liu2020enhancing,gutoski2021comparative}. While this technique is ubiquitous in video understanding, prior approaches only consider the transfer from one single source task (dataset) and are thus unable to model the relations among multiple video tasks simultaneously. 

Taskonomy~\cite{zamir2018taskonomy} presents task transfer with a thorough analysis on the structure of multiple visual tasks. Many works~\cite{pal2019zero,standley2020tasks,dwivedi2019representation,zamir2020robust} continue along this direction and explore visual task relations, yet they limit the discussion to static images and generally require a unified design across all tasks. In contrast, we consider a diverse set of egocentric video tasks, which are addressed with a heterogeneous set of task-specific video architectures 
(\eg, accommodating different time, space, or multimodality). Clearly, forcing the same network architecture across all tasks can be suboptimal for each individual task. This motivates our proposed \ett-s (Figure \ref{fig:related}(c)), where we preserve the heterogeneous backbones developed for each task and build a task translator on top of the task-specific models.



\textbf{Multi-task Learning}. In MTL~\cite{mtlsurvey}, a single model is trained to address multiple tasks simultaneously in order to capture synergistic supervision across tasks. As depicted in Figure \ref{fig:related}(b), hard parameter sharing~\cite{mtloverview} (\ie, sharing a backbone among tasks and keeping one separate head for each task) is the most commonly used technique within this genre. Although MTL has shown to be beneficial of video analysis~\cite{ma2016going,luvizon2020multi,fathi2012learning,li2018eye,huang2020mutual,baradel2018object,kapidis2019multitask}, there is ongoing debate about the best strategies to determine what parameters to share across which tasks~\cite{kang2011learning,chen2017divide,standley2020tasks,guo2020learning,sun2020adashare}. As pointed out in ~\cite{kokkinos2017ubernet}, when MTL is achieved by means of a single common backbone, the performance tends to decrease when the number of tasks grows beyond a certain point. Furthermore, many works~\cite{gong2019comparison,leang2020dynamic,standley2020tasks,guo2020learning} observe that over-sharing a network across unrelated tasks causes negative transfer and hinders individual task performance. 
While soft parameter sharing~\cite{duong2015low,yang2016trace} mitigates this by retaining distinct copies of parameters, it still requires adopting the same identical architecture  and ``similar'' weight values across all tasks.

%
In the video domain, several works utilize synergies between related tasks (\eg, action recognition with gaze prediction ~\cite{fathi2012learning,li2018eye,huang2020mutual} or body pose estimation~\cite{luvizon2020multi}). \revision{However, when selected tasks are not strongly related, prior approaches that split the learning capacity of a shared backbone over multiple tasks can suffer from task competition and inferior performance}. In the image domain, with the great advancement of transformers~\cite{vaswani2017attention}, training with multiple datasets together for a generalist model is gaining popularity. Recent work~\cite{chen2022unified,kolesnikov2022uvim,lu2022unified,girdhar2022omnivore,jaegle2021perceiver,jaegle2022perceiver} investigates a unified transformer architecture across a diverse set of tasks. Our variant \ett-g (Figure \ref{fig:related}(d)) is motivated by the desiderata of shared knowledge encapsulated by MTL and of a generalist model. Unlike previous learning paradigms, we adopt a ``flipped design'' involving separate task-specific backbones and a task translator shared across all tasks. This effectively mitigates task competition and achieves task translation for all tasks simultaneously. 

\section{Approach}
We are given $K$ video tasks, ${\mathcal T_k}$ for $k=1, \cdots, K$. We note that our approach does not require a common training set with annotations for all tasks. 
Let the dataset for task $\mathcal T_k$ be $\mathcal D
^{\mathcal T_k} = \{(\mathbf x_i^{\mathcal T_k}, y_i^{\mathcal T_k})\}_{i=1}^{N_k}$, where $(\mathbf x_i^{\mathcal T_k}, y_i^{\mathcal T_k})$ denotes the $i$-th pair of (input video, output label) and $N_k$ represents the number of given examples. Note that ``labels" $y_i^{\mathcal T_k}$ can be a variety of output types, and are not limited to category labels. For simplicity we omit the subscript $i$ hereafter.  

We consider two formulations 
with distinct advantages: (1) task-specific translation, where we partition the tasks into one primary task $\mathcal T_p$ and $K-1$ auxiliary tasks, and optimize the objective to improve performance on $\mathcal T_p$ with the assistance of the auxiliary tasks (\ett-s, Sec.~\ref{sec:method_ts}); (2) task-general translation, where we treat all $K$ tasks equally, and the goal is to maximize the collective performance of all the tasks (\ett-g, Sec.~\ref{sec:method_tg}). As demonstrated in our experiments, objective (1) leads to the strongest performance on the primary task, while objective (2) offers the benefit of a single unified model addressing all tasks at once. 


\subsection{Task-Specific Translation: EgoT2-s}\label{sec:method_ts}

\begin{figure*}[!t]
  \centering
  \includegraphics[width=1.0\linewidth]{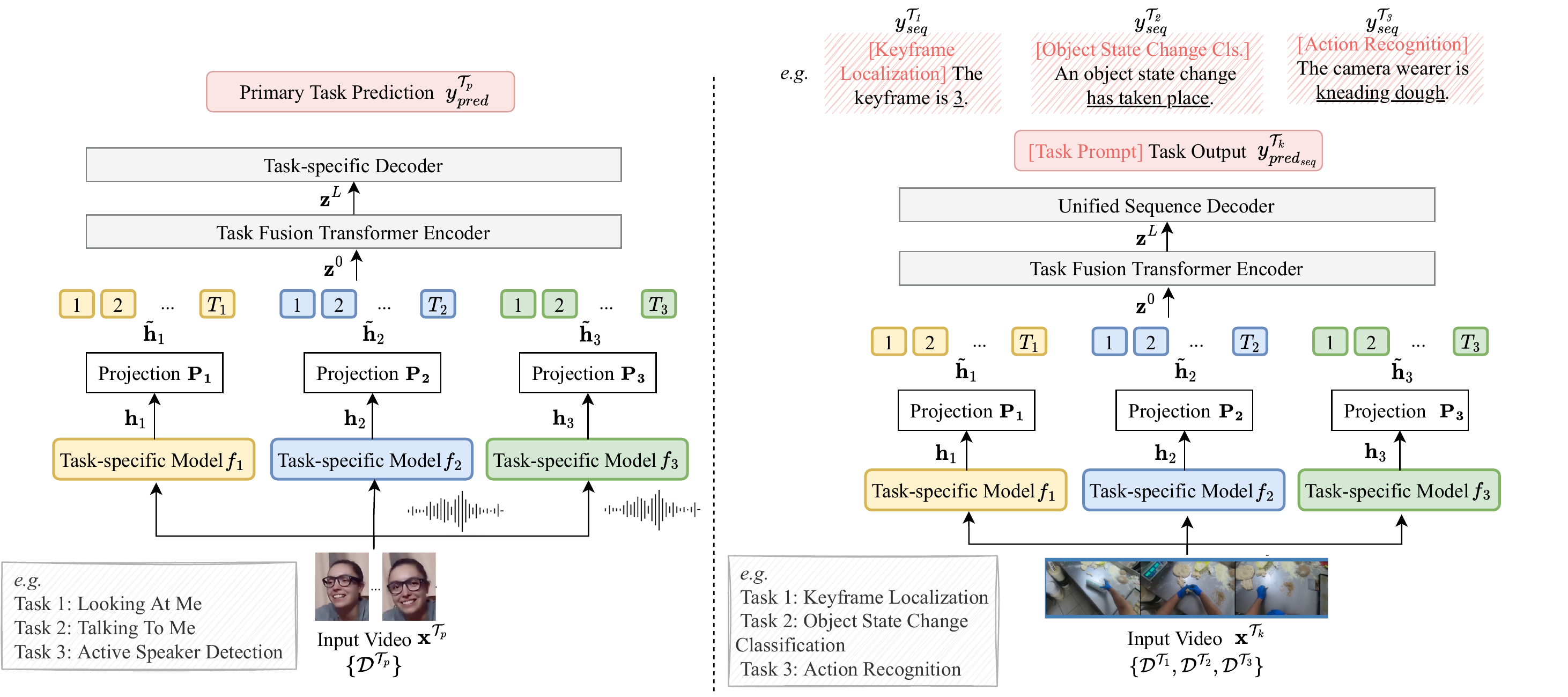}
  \caption{An illustration of \ett-s (left) and \ett-g (right) on three candidate tasks. The left figure illustrates \ett-s~on three social interaction tasks, where the input to each model is unimodal (\ie, video) or multimodal (\ie, video and audio). The right figure shows the design of \ett-g on three example tasks that focus on different aspects of human-object interactions (\ie, localization, object state change classification, and action recognition). \revision{\ett-s learns to ``translate'' auxiliary task features into predictions for the primary task and \ett-g conducts task translation conditioned on the task of interest.} 
  }
  \label{fig:framework}
\end{figure*}

The training of \ett-s is split over two stages. 

{\bf \noindent Stage I: Individual-Task Training.} We train a separate model $f_k$ on each individual task dataset $\mathcal D^{\mathcal T_k}$, obtaining $K$ task-specific models $\{f_k\}_{k=1}^K$. 
We do not place any restrictions on the task-specific model designs, nor do we require a unified design (\ie, identical encoder-decoder architecture)  
across tasks.
Therefore, any available model checkpoint developed for task $\mathcal T_k$ can be adopted as $f_k$ within our framework, offering maximum flexibility.

{\bf \noindent Stage II: Task-Specific Translation.}  We train a task translator that takes features produced by task-specific models as input and outputs predictions for the primary task. Formally, let $\mathbf h_k \in \mathbb{R}^{T_k \times D_k}$ be features produced by the $k$-th task-specific model $f_k$, where $T_k$ is the temporal dimension and $D_k$ is the per-frame feature dimension for model $f_k$. Following the feature extraction step, we design a projection layer $\mathbf P_k\in \mathbb{R}^{D_k \times D}$ for each $f_k$ to map task-specific features to a shared latent feature space. This yields a temporal sequence of task-specific tokens $\tilde{\mathbf h}_k \in \mathbb{R}^{T_k \times D}$. 

We process this collection of task-specific temporal sequences using a transformer encoder~\cite{vaswani2017attention} of $L$ layers to capture both \emph{inter-frame} and \emph{inter-task} dependencies. We denote the propagation rule of layer $l$ by $\mathbf z^{l+1} = Encoder^{l}(\mathbf z^l)$. Finally, we adopt a decoder head $Decoder^{\mathcal T_p}$ to obtain predictions for the primary task $\mathcal T_p$. 

In all, this stage has four major steps: (1) feature extraction; (2) feature projection; (3) transformer fusion; and (4) feature decoding. The procedure is summarized below:
\begin{align}
    \mathbf h_k & = f_k(\mathbf x^{\mathcal T_p}),\quad \forall k\in \{1,2,\cdots, K\} \label{eq:feat_extract}\\
    \tilde{\mathbf h}_k & = \mathbf P_k \mathbf h_k, \quad \forall k\in \{1,2,\cdots, K\}\\
    \begin{split}
    \mathbf z^0 & = [\tilde{\mathbf h}_1, \tilde{\mathbf h}_2, \cdots, \tilde{\mathbf h}_K] \\
    \mathbf z^{l+1} & = Encoder^{l}(\mathbf z^l), \forall l\in \{0,1,\cdots, L-1\} 
    \end{split} \\
    y_{pred}^{\mathcal T_p} & = Decoder^{\mathcal T_p}(\mathbf z^L) \label{eq:decode}
\end{align}
where $y_{pred}^{\mathcal T_p}$ denotes the prediction given by \ett-s. During the second stage of training, we freeze the task-specific models and optimize the task translator with respect to the primary task dataset $\mathcal D^{\mathcal T_p}$. 

Figure \ref{fig:framework} (left) 
illustrates the design of \ett-s using three social interaction tasks from Ego4D~\cite{ego4d} as an example. \ett-s allows heterogeneity in the task-specific models (\ie, $f_1$ is unimodal while $f_2$ and $f_3$ are multimodal; also the three task-specific models can be associated with different frame rates and temporal durations) and utilizes a transformer encoder to model inter-frame and inter-task relations. The resulting \ett-s learns to adaptively utilize auxiliary task features for the primary task prediction. 

\subsection{Task-General Translation: EgoT2-g}\label{sec:method_tg}

\ett-s optimizes performance for a single primary task. Therefore, in the event all $K$ tasks must be addressed, it requires $K$ separate training runs and $K$ distinct translators. This motivates us to extend \ett-s to perform task translation for all $K$ tasks at once. In \ett-g, the task translator processes features from all $K$ tasks and learns to ``translate'' features conditioned on the task of interest. 

The first stage of \ett-g is identical to \ett-s. For the second stage, we propose two main modifications. First, we replace the task-specific decoder in \ett-s with a ``generalist'' decoder that outputs predictions conditioned on the task of interest. Natural language provides us with a flexible scheme to specify all tasks as a sequence of symbols. Inspired by~\cite{chen2022unified}, we tokenize all task outputs and replace the original task-specific decoder with a sequence decoder~\cite{radford2019language} for a unified interface. Specifically, we first transform the original label $y^{\mathcal T_k}$ to a target output sequence $\mathbf{y}_{seq}^{\mathcal T_k} \in \mathbb R^{M}$, where $M$ is the target sequence length. For the task translator to produce task-dependent outputs, we prepend a task prompt token $\mathbf{y}_{prompt}$ to the target output, 
\ie, $\mathbf{y}_{seq_1}^{\mathcal T_k} = \mathbf{y}_{prompt}$. We then let the sequence decoder generate a sentence answering the requested task. 
Figure \ref{fig:framework} (right) illustrates how we express task outputs as sequences of discrete tokens and attach task prompts. 

With the transformed output, we treat the problem as a language modeling task and train the task translator to predict subsequent tokens (one token at a time) conditioned on the input video and its preceding tokens. The training objective is $\mathcal L^{\mathcal T_k} = \sum_{j=1}^M \mathbf w_j \log P(\mathbf y_{seq_j}^{\mathcal T_k} | \mathbf{x}^{\mathcal T_k}, \mathbf y_{seq_{1:j-1}}^{\mathcal T_k})$. 
Note that the maximum likelihood loss is weighted to mask the loss corresponding to the task prompt token: $\mathbf w_j$ is set to 0 for $j=1$, and to 1 for any other $j$. During inference, the task prompt is prepended, and the task translator predicts the remaining output tokens. We use argmax sampling (\ie, take the token with the largest likelihood) to sample tokens from the model likelihood and transform the output tokens back to the original label space. Detokenization is easy as we simply reverse the tokenization process. 


The second modification lies in the training strategy. While \ett-s adopts the primary task dataset for training, \ett-g requires joint training on all $K$ task datasets. Similar to the training strategy in ~\cite{girdhar2022omnivore,chen2022unified}, we sample one batch from each task, compute the task loss, aggregate the $K$ gradients, and perform model updates in one training iteration. 
The final training objective is $\mathcal L = \sum_{k=1}^K \mathcal L^{\mathcal T_k}$. 

Figure \ref{fig:framework} contrasts the design of \ett-s and \ett-g. They both provide a flexible framework that can incorporate multiple heterogeneous task-specific models (\eg, the three example tasks we give here focus on different aspects of human-object interactions). With a design and an optimization that are specialized to a single primary task, \ett-s is expected to lead to superior individual task performance while \ett-g brings the efficiency and compactness benefits of a single translator addressing all tasks. 

\section{Experiments}\label{sec:exp}


\subsection{Experimental Setup}\label{sec:exp_setup}
\begin{figure}[!t]
  \centering
  \includegraphics[width=0.8\linewidth]{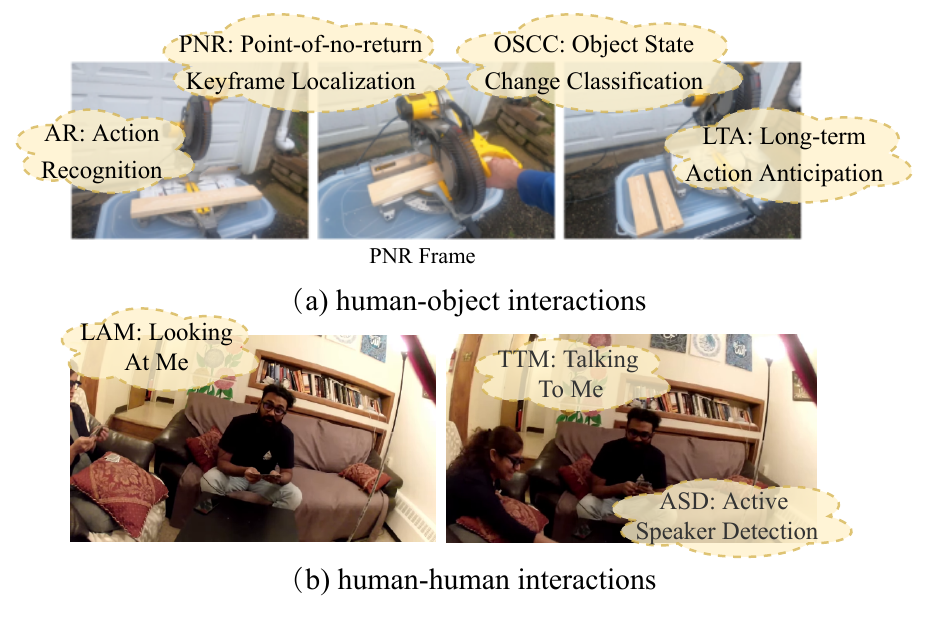}
  \vspace*{-5mm}
  \caption{Task Setup. We select a broad set of egocentric video tasks that focus on (a) human-object interactions and (b) human-human interactions from Ego4D benchmarks.}
  \label{fig:task}
\end{figure}

{\bf \noindent Dataset and Tasks.}  We evaluate on Ego4D~\cite{ego4d}, the world's largest egocentric dataset with 3,670 hours of videos spanning hundreds of scenarios (\eg, household, outdoor, leisure). It offers five benchmarks: 
episodic memory (EM), hands and objects (HO), audio-visual diarization (AV), social interactions (Social) and forecasting. For our study, we select 7 tasks spanning 4 benchmarks, representing a variety of tasks in egocentric perception, \revision{as illustrated in Figure \ref{fig:task}}. The 7 tasks fall into two broad clusters: (a) human-object interactions and (b) human-human interactions. Table \ref{tab:dataset} summarizes our task setup. For each cluster, we use tasks from the same benchmark as well as tasks across benchmarks, in an attempt to reveal connections among seemingly unrelated tasks. 
The 7 candidate tasks are heterogeneous in nature as they are defined on videos of varying duration, adopt different video models as backbones, and process unimodal (\ie, video) or multimodal (\ie, video and audio) input, offering a diverse task setup for our study. See Appendix \ref{ap_sec:data} for more details. 

{\bf \noindent Models and Baselines.}  For each task, we adopt for consistency the baseline models introduced with the Ego4D dataset\footnote{We use model checkpoints provided on the Ego4D website: \url{https://github.com/EGO4D}.} as the task-specific \textbf{(TS) models} in \ett. For task-specific translation (Sec.~\ref{sec:exp_ts}), we train one task translator for each primary task and use all the other tasks in the same cluster (either human-object interactions or human-human interactions) as auxiliary tasks.
We compare \ett-s with two representative transfer learning approaches: (1) \textbf{Transfer}~\cite{zamir2018taskonomy} denotes finetuning a transfer function on top of features produced by the auxiliary task models (Figure \ref{fig:related}(a)). (2) \textbf{Late Fusion~\cite{ma2016going} (LF)}  concatenates auxiliary task features along with primary task features, and finetunes a few layers that receive the concatenated features as input for the final prediction. Furthermore, to gauge possible improvements over TS by increasing capacity, we consider a \textbf{Finetuning}~\cite{donahue2014decaf} baseline, which finetunes a few layers on top of the features produced by the primary task model. In order to make a fair comparison, the first-stage training of these baselines is identical to that of \ett, and the number of parameters in the second stage of training is set to match that of \ett-s as closely as possible. 

\begin{table}[!t]
  \centering
  \scalebox{0.73}{
  \begin{tabular}{llcccc}
    \toprule
   & Task & Benchmark & Mod. & \makecell[c]{Duration\\ (seconds)} & \makecell[c]{Model \\backbone}\\
    \midrule
 \multirow{4}{*}{(a)} & PNR & HO & V & 8.0 & I3D RN-50~\cite{carreira2017quo} \\
  &  OSCC & HO & V & 8.0 & I3D RN-50~\cite{carreira2017quo} \\
  &  AR & Forecasting & V & 8.0 & SlowFast~\cite{feichtenhofer2019slowfast} \\
  &  LTA & Forecasting & V & 16.0 & SlowFast~\cite{feichtenhofer2019slowfast} \\
  \midrule
  \multirow{3}{*}{(b)} & LAM & Social & V & 0.2 & 3D RN-18~\cite{tran2015learning} \\ 
   & TTM & Social & A\&V & 2.7 & 3D RN-18~\cite{tran2015learning} \\
   & ASD & AV & A\&V & 3.7 & TalkNet  ~\cite{tao2021someone} \\
    \bottomrule
  \end{tabular}}
  \caption{Task Descriptions. `Mod.' is short for modality; `A' and `V' denote audio and video, respectively. }
  \label{tab:dataset}
\end{table}

For task-general translation (Sec.~\ref{sec:exp_tg}), the task translator is jointly optimized for all tasks within a cluster\footnote{There is a significant domain gap between human-human and human-object interaction videos. See Appendix \ref{ap_sec:result} for cross-cluster \ett-g.}
, thus we have one task translator for human-object interactions that attends to all tasks simultaneously and one translator that performs three human-human interaction tasks at the same time. 
For comparison with \ett-g, we implement the most widely adopted \textbf{multi-task} learning approach, hard parameter sharing~\cite{mtloverview} (Figure \ref{fig:related}(b)). 

\begin{table*}[t]
  \centering
  \scalebox{0.8}{
  \begin{tabular}{lcc|cc|ccc|ccc}
    \toprule
     & \multicolumn{2}{c}{$\mathcal T_p$ is PNR} & \multicolumn{2}{c}{$\mathcal T_p$ is OSCC} & \multicolumn{3}{c}{$\mathcal T_p$ is AR} & \multicolumn{3}{c}{$\mathcal T_p$ is LTA}   \\
     & \# Params $\cdot 10^6$ & Error  & \# Params $\cdot 10^6$ & Acc. & \# Params $\cdot 10^6$ & \multicolumn{2}{c}{Acc. (\%) $\uparrow$} & \# Params $\cdot 10^6$ & \multicolumn{2}{c}{ED@20 $\downarrow$} \\
     & Trainable (\emph{All}) & (s) $\downarrow$ & Trainable (\emph{All}) & (\%) $\uparrow$ & Trainable (\emph{All}) & Verb & Noun & Trainable (\emph{All}) & Verb & Noun \\
    \midrule
  TS model ~\cite{ego4d} & 32.2 \emph{(32.2)} & 0.615 & 32.2 \emph{(32.2)} & 68.22 & 63.3 \emph{(63.3)} & 22.18 & 21.55 & 180 \emph{(242)} & 0.746 & 0.789 \\
    \midrule
Finetuning~\cite{donahue2014decaf} & 8.4 \emph{(40.6)} & 0.611 & 8.4 \emph{(40.6)} & 67.93 & 4.9 \emph{(66.8)} & 21.64 & 22.84 & 48.6 \emph{(266)} & 0.744 & 0.787 \\
    Transfer~\cite{zamir2018taskonomy} (PNR) & N/A & N/A & 8.4 \emph{(40.6)} & 66.80 & 4.9 \emph{(37.1)} & 19.98 & 5.44 & 65.4 \emph{(97.6)} & 0.778 & 0.902 \\
   Transfer~\cite{zamir2018taskonomy} (OSCC) & 8.4 \emph{(40.6)} & 0.611 & N/A & N/A & 4.9 \emph{(37.1)} & 20.00 & 9.61 & 65.4 \emph{(97.6)} & 0.774 & 0.899  \\
 Transfer~\cite{zamir2018taskonomy} (AR) & 9.5 \emph{(71.4)} & 0.613 & 9.4 \emph{(71.4)} & 70.98 & N/A & N/A & N/A & 53.3 \emph{(115)} & 0.745 & 0.806\\
 LF~\cite{ma2016going} (All Tasks) & 9.6 \emph{(135)} & 0.610 & 9.6 \emph{(135)} & 72.10 & 5.2 \emph{(131)} & 21.11 & 19.24 & 83.6 \emph{(427)} & 0.744 & 0.788 \\ 
 \ett-s (All Tasks) & 6.4 \emph{(132)} & \textbf{0.610} & 7.4 \emph{(133)} & \textbf{72.69} & 4.3 \emph{(130)} & \textbf{23.04} & \textbf{23.28} & 41.8 \emph{(348)} & \textbf{0.731} & \textbf{0.769} \\  
    \bottomrule
  \end{tabular}
 }
  \caption{Results of \ett-s as we vary the primary human-object interaction task $\mathcal T_p$. 
  First row records performance of the task-specific (TS) model we obtain in the first-stage training; we compare \ett-s with other baseline methods in the second-stage training. 
  We list the number of trainable parameters for each separate stage as well as the total (\ie, trainable parameters plus parameters of frozen TS models) in parentheses. Following~\cite{ego4d}, the evaluation metric is temporal localization error (unit: seconds) for PNR, accuracy for OSCC and AR, and edit distance at future 20 time stamps (\ie, ED@20) for LTA. For localization error and ED@20, lower is better.  \ett-s reliably adapts the auxiliary tasks to suit the target task. 
  }
  \label{tab:ts_HOI}
\end{table*}

\begin{table}[h]
  \centering
   \resizebox{1.0\columnwidth}{!}{
  \begin{tabular}{lcc|cc}
    \toprule
  & \multicolumn{2}{c}{$\mathcal T_p$ is TTM} & \multicolumn{2}{c}{$\mathcal T_p$ is ASD} \\
     & \makecell[c]{\# Params $\cdot 10^6$ \\ Trainable (\emph{All})} &  \makecell[c]{mAP  \\ (\%) $\uparrow$} & \makecell[c]{\# Params $\cdot 10^6$ \\ Trainable (\emph{All})} &  \makecell[c]{mAP \\ (\%) $\uparrow$} \\
    \midrule
  TS model~\cite{ego4d} & 20.2  \emph{(20.2)} & 58.91 & 15.7 \emph{(15.7)} & 79.05\\
 \midrule 
 Finetuning~\cite{donahue2014decaf} & 0.8 \emph{(20.8)} & 59.67 & 1.1 \emph{(16.8)} & 78.62  \\
Transfer~\cite{zamir2018taskonomy} (LAM) & 0.8 \emph{(15.4)} & 63.59 & 1.6 \emph{(16.2)} & 66.40 \\
Transfer~\cite{zamir2018taskonomy} (TTM) & N/A & N/A & 1.6 \emph{(21.6)} & 71.06  \\
 Transfer~\cite{zamir2018taskonomy} (ASD) & 0.8 \emph{(16.5)} & 62.31 & N/A & N/A \\
 LF~\cite{ma2016going} (All Tasks) & 1.2 \emph{(51.5)} & 64.29 & 1.6 \emph{(51.9)} & 77.54\\
 \ett-s (All Tasks) &0.7 \emph{(51.1)}& \textbf{66.54} & 1.5 \emph{(51.9)}& \textbf{79.38} \\
 \bottomrule
  \end{tabular}
}
  \caption{Results of \ett-s 
  as we vary the primary human-human interaction task $\mathcal T_p$. \ett-s consistently improves the TS model.
  }
  \label{tab:ts_HHI}
\end{table}

{\bf \noindent Implementation Details.}
\revision{There is one video preprocessing step before the feature extraction step in Equation (\ref{eq:feat_extract}), where we transform the original video input from $\mathbf x^{\mathcal T_p}$ to match the input format of the $k$-th task-specific model $f_k$. In particular, $\mathbf x^{\mathcal T_p}$ is first upsampled or downsampled to match the frame rates required by $f_k$. Next, if the temporal span of the auxiliary task is smaller than that of the primary task, we slide $f_k$ in a moving window to extract a sequence of features, where the window length is the temporal span required by $f_k$, and stride size is a hyperparameter. Conversely, if $f_k$ requires video inputs of a longer temporal span than $\mathbf x^{\mathcal T_p}$, \revision{we exclude task $k$ from auxiliary task candidates to avoid providing potential advantages of a longer observation window to our framework as otherwise we need to expand video length of $\mathbf x^{\mathcal T_p}$ to match the requirement of $f_k$.} 
Moreover, 
if the auxiliary task dataset is multimodal (\ie, video and audio) and the primary task involves only video, we apply the unimodal video pathway of $f_k$ to obtain features; 
if the primary task is multimodal, we provide all task-specific features that are computable from these modalities. See Appendix \ref{ap_sec:implem} for more implementation details.} 


\subsection{Evaluation of Task-Specific Translation} \label{sec:exp_ts}

{\bf \noindent Results.} We conduct experiments with \ett-s for each task being the primary task\footnote{Following the time-span guidelines in Sec.~\ref{sec:exp_setup}, LAM is not considered as the primary task and LTA is not adopted as an auxiliary task. 
Nonetheless, Appendix \ref{ap_sec:result} shows some special cases for completeness.} and summarize the results for human-object interactions and human-human interactions in Table \ref{tab:ts_HOI} and \ref{tab:ts_HHI}, respectively. 

From the two tables, we observe uneven performance by the baseline methods. Transfer and Late Fusion sometimes outperform the dedicated TS model and sometimes underperform it. 
When tasks do not exhibit a strong transfer relation, reusing the backbone of the auxiliary task for the primary task leads to negative transfer and performance degradation. For instance, in Table \ref{tab:ts_HOI}, when $\mathcal T_p$ is AR,
Transfer (OSCC) and Late Fusion both downgrade noun prediction accuracy, suggesting 
object state change is more dependent on verbs and unrelated to noun prediction tasks in AR. 

On the contrary, our proposed \ett-s learns to adaptively utilize task-specific features and effectively mitigates negative transfer, demonstrating consistent improvement over the TS model for all 6 cases. For instance, in Table \ref{tab:ts_HHI}, when $\mathcal T_p$ is ASD, Late Fusion indicates there is a deleterious relation from LAM and TTM to ASD, as it suffers from an accuracy degradation of 1.51\% over TS, yet \ett-s still obtains slightly better performance compared to TS (\ie, 79.38\% v.s. 79.05\%). Moreover, when auxiliary tasks are beneficial for the primary task, \ett-s outperforms all baselines with fewer trainable parameters. For example, when $\mathcal T_p$ is TTM, it achieves a +7.63\% mAP improvement over the original TS model by training a lightweight task translator with only 0.7M parameters on top of it (TS is kept frozen). These results across different primary and auxiliary task combinations help demonstrate 
 the generalizability of \ett-s. 
 See Appendix~\ref{ap_sec:result} for experiments using a subset of auxiliary tasks rather than all tasks. 

{\bf \noindent Ablation Study.} In Table \ref{tab:ablation}, we ablate three different design choices of \ett-s using TTM as the primary task: (a) We replace the LAM and ASD TS models in \ett-s with two TTM models with different parameters. This yields a task fusion transformer that is architecturally identical to \ett-s but takes only TTM tokens as input; (b) We pass features produced by TS models after temporal pooling as the input of our task fusion transformer; (c) We do not freeze TS models in our second-stage training. By comparing (a) with our default configuration (d), we see that \ett-s indeed benefits from the introduction of auxiliary tasks. Although equipped with three different TTM models and a larger model size (the total number of parameters of three TTM models is larger than the sum of three TS models), variant (a) does not bring as much performance gain as \ett-s (d). Also, preserving the temporal information of task-specific tokens further boosts performance, as can be seen in the comparison of \ett-s (b) with \ett-s (d). Finally, 
not freezing TS (c) greatly increases the training cost yet brings no performance gain. These results validate the design of our proposed \ett-s. 
\begin{table}[!t]
 \centering
 \scalebox{0.8}{
  \begin{tabular}{lccccc}
  \toprule
  & \makecell[c]{\# Params $\cdot 10^6$ \\ Trainable (\emph{All})} & \makecell[c]{Auxiliary\\ Tasks} & \makecell[c]{Temporal\\ Information} & \makecell[c]{Frozen\\ TS model} & \makecell[c]{mAP \\(\%) $\uparrow$} \\
  \midrule
  (a) & 0.7 (\emph{60.8}) & & $\checkmark$ & $\checkmark$ & 63.40\\
  (b) & 0.7 (\emph{51.1})& $\checkmark$ &  & $\checkmark$ & 65.47 \\
  (c) & 51.1 (\emph{51.1}) & $\checkmark$ & $\checkmark$ & & 66.00 \\
  (d) & 0.7 (\emph{51.1})& $\checkmark$ & $\checkmark$ & $\checkmark$ & 66.54\\
  \bottomrule
  \end{tabular}
}
\caption{Ablation study of \ett-s ($\mathcal{T}_p$ is TTM). 
}
  \label{tab:ablation} 
\end{table}


\subsection{Evaluation of Task-General Translation} \label{sec:exp_tg}

\setlength\tabcolsep{2pt}
\begin{table}[!t]
  \centering
\resizebox{1.0\columnwidth}{!}{
  \begin{tabular}{lccccccc}
    \toprule
    \makecell[c]{(a)} & \makecell[c]{\# Params \\Trainable} & \makecell[c]{PNR\\$\downarrow$} & \makecell[c]{OSCC\\$\uparrow$} & \makecell[c]{AR\\ Verb $\uparrow$ }& \makecell[c]{AR\\ Noun $\uparrow$} & \makecell[c]{LTA\\ Verb $\uparrow$} & \makecell[c]{LTA\\ Noun $\uparrow$}\\
     \midrule
     TS model~\cite{ego4d} & N/A & 0.615 & 68.2 & \bf{22.18} & 21.55 & 20.82 & 21.80\\
     Multi-task~\cite{mtloverview} & 32.2 & 0.617 & 66.0 & N/A & N/A & N/A & N/A \\
     \hdashline
     \makecell[l]{\ett-g (P \& O)} & 5.9 & 0.612 & 68.6 & N/A & N/A & N/A & N/A \\
      \makecell[l]{\ett-g (All)} & 34.5 & \bf{0.611} & \bf{71.7} & 21.93 & \bf{22.73} & \bf{21.91} & \bf{23.61} \\
    \midrule
  \end{tabular}
}
  \label{tab:HOI}
\end{table}

\setlength\tabcolsep{6pt}
\begin{table}[!t]
  \centering
  \scalebox{0.8}{
  \begin{tabular}{lcccc}
    \midrule
    \makecell[c]{(b)} & \makecell[c]{\# Params \\Trainable} & \makecell[c]{LAM\\ mAP (\%) $\uparrow$} & \makecell[c]{TTM\\ mAP (\%) $\uparrow$} & \makecell[c]{ASD\\ Acc.(\%) $\uparrow$} \\
     \midrule
    TS model~\cite{ego4d} & N/A & \bf{77.79} & 58.91 & 79.05\\
    Multi-task~\cite{mtloverview} & 20.2 & 60.53 & 61.91 & N/A \\
    \hdashline
   \ett-g & 1.4 & 77.63 & \bf{64.49} & \bf{79.06} \\
    \bottomrule
  \end{tabular}
}
  \caption{\ett-g for (a) human-object interaction and (b) human-human interaction tasks. The evaluation metric is error (seconds) for PNR (P) and accuracy (\%) for OSCC (O), AR and LTA. 
  We report the number of trainable parameters required for each method in the second-stage training (unit: million). Our model is flexible, accurate, and avoids negative transfer.}
  \label{tab:tg}
\end{table}

{\bf \noindent Results}. Table \ref{tab:tg} provides results of \ett-g. 
Since the TTM and LAM baseline models use identical video backbones (\ie, 3D ResNet-18), the hard parameter sharing multi-task baseline~\cite{mtloverview} can jointly learn TTM and LAM. Yet this model design is unable to solve the ASD task without further modifications to the ASD backbone model. In contrast, our \ett-g provides a flexible solution that can incorporate a heterogeneous mix of pretrained models. Similarly, we apply the multi-task baseline to PNR and OSCC, as they use the same video backbone (\ie, I3D ResNet-50). Compared with dedicated TS models, our proposed \ett-g performs task translation for \emph{all} tasks at the same time and achieves on parallel or better performance for all tasks. For instance, it achieves +5.58\% mAP improvement for TTM and 3.5\% accuracy gain for OSCC. Notably, on ASD, it retains the top-performance of the original TS models when the other two auxiliary tasks do not help. In contrast, we observe task competition for the multi-task baseline: the improvement for TTM (\ie, +3.0\% mAP) is at the cost of significantly downgraded LAM performance (\ie, -17.26\% mAP). Similarly, sharing an encoder for PNR and OSCC also leads to task competition and suboptimal performance for the multi-task baseline. \revision{For a side-by-side comparison, we also implement \ett-g that performs task translation for PNR and OSCC only and observe its advantages over the multi-task baseline in terms of both performance and trainable parameters. As \ett-g does not require re-training of the backbone, we can integrate any available model checkpoint developed for each individual task into our framework and train a lightweight task-general translator to further boost performance in the second stage.}

\begin{table}[!t]
 \centering
  \scalebox{0.8}{
  \begin{tabular}{lccc}
  \toprule
  \textit{TTM Challenge} & \multicolumn{3}{c}{mAP $\uparrow$}\\
  \hdashline
 Random Guess~\cite{ego4d} & \multicolumn{3}{c}{0.50} \\
 3D ResNet-18 Bi-LSTM~\cite{ego4d} & \multicolumn{3}{c}{0.54}\\
  \ett-g (3D ResNet-18) & \multicolumn{3}{c}{0.58} \\
 \ett-s (3D ResNet-18) & \multicolumn{3}{c}{\textbf{0.58}} \\
    \midrule
\textit{PNR Challenge} & \multicolumn{3}{c}{Error (s) $\downarrow$}\\
\hdashline
   Always Center Frame~\cite{ego4d} &  \multicolumn{3}{c}{1.01} \\
   CNN LSTM~\cite{ego4d} &  \multicolumn{3}{c}{0.76}\\
   EgoVLP~\cite{lin2022egocentric} &  \multicolumn{3}{c}{0.67} \\
   Video Swin Transformer~\cite{escobar2022video} &  \multicolumn{3}{c}{0.66} \\
   SViT~\cite{ben2022structured} &  \multicolumn{3}{c}{0.66} \\
   \ett-s (I3D ResNet-50)&  \multicolumn{3}{c}{\textbf{0.66}} \\
  \midrule
  \textit{OSCC Challenge} & \multicolumn{3}{c}{Acc. $\uparrow$} \\
\hdashline
   Always Positive~\cite{ego4d} & \multicolumn{3}{c}{0.48} \\ 
   I3D ResNet-50~\cite{ego4d} &\multicolumn{3}{c}{0.68} \\ 
   Video Swin Transformer~\cite{escobar2022video} &  \multicolumn{3}{c}{0.68} \\
   Divided ST Attention~\cite{islam2022object} & \multicolumn{3}{c}{0.72}\\
   EgoVLP~\cite{lin2022egocentric} & \multicolumn{3}{c}{0.74} \\
   \ett-g (I3D ResNet-50) & \multicolumn{3}{c}{0.70} \\
   \ett-s (I3D ResNet-50) & \multicolumn{3}{c}{0.71} \\
   \ett-s (EgoVLP) & \multicolumn{3}{c}{\textbf{0.75}} \\
  \midrule
  \multirow{2}{*}{\textit{LTA Challenge}} & \multicolumn{3}{c}{ED@20 $\downarrow$}\\
   & Verb & Noun & Action \\
  \hdashline
  SlowFast + Transformer~\cite{ego4d} & 0.74 & 0.78 & 0.94\\
   Video + CLIP ~\cite{das2022video+} & 0.74 & 0.77 & 0.94 \\
   Hierarchical MLP Mixer~\cite{mascaro2022intention} & 0.74 & \textbf{0.74} & \textbf{0.93} \\
  \ett-s (SlowFast)& \textbf{0.72} & 0.76 & \textbf{0.93} \\
  \bottomrule
  \end{tabular}
}
\caption{Comparison of \ett \ with SOTA approaches on four Ego4D challenges (test set). We list the TS model architecture of \ett \ in parentheses.  Our model improves the state of the art.
}
  \label{tab:cmp_sota} 
\end{table}

{\bf \noindent Comparison with SOTA Approaches.} To further demonstrate the efficacy of both \ett-s and \ett-g, we submit our model to the EvalAI server to compare it with winning submissions to Ego4D-CVPR'22 and Ego4D-ECCV'22 challenges on the withheld test set. Table \ref{tab:cmp_sota} shows the results.\footnote{ASD \& AR are not applicable since they are not Ego4D challenges. Results of \ett-g for PNR \& LTA are unavailable (see Appendix \ref{ap_sec:implem}).} 
\ett-s achieves top performance for all 4 challenges. By only incorporating basic video backbones (\eg, 3D ResNet-18 and SlowFast) as the task-specific model, \ett-s achieves similar or better performance than works that adopt more powerful, novel architectures such as Video Swin Transformer. Moreover, 
the benefits of our approach are orthogonal to such architecture improvements: \eg, for the OSCC challenge, replacing the I3D ResNet-50 backbone with the one used in EgoVLP~\cite{lin2022egocentric} can further elevate the accuracy of EgoVLP by 1\%. This indicates the success of \ett \ stems from its effective use of task synergies.

While \ett-g is a strong performer that surpasses or matches TS across all tasks, if we compare its results
with those of \ett-s, 
we observe that \ett-s demonstrates superior performance.  
This is understandable given that \ett-s is individually optimized for each primary task and employs a specialized translator. On the other hand, \ett-g provides a favorable unified framework that performs task translation for all tasks simultaneously via the design of a task-general translator. Thus, \ett-s serves as the framework of choice for top performance while \ett-g provides added flexibility. See Appendix \ref{ap_sec:result} for a detailed comparison of the performance and efficiency of these two variants.




\subsection{Visualization of Uncovered Task Relations}

\begin{figure}[!t]
  \centering
  \includegraphics[width=1.0\linewidth]{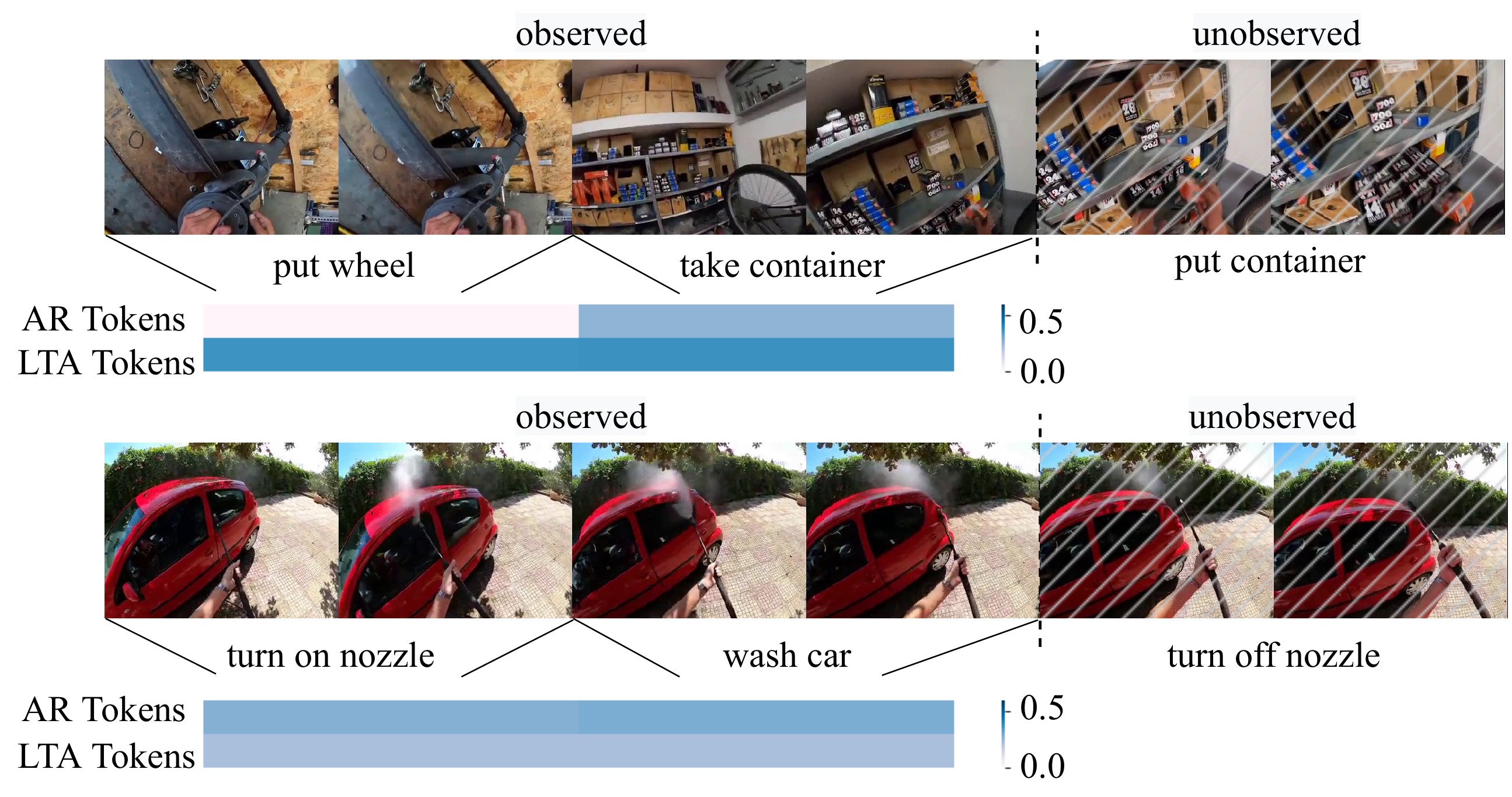}
  \caption{Attention weights of \ett-s when $\mathcal{T}_p$ is LTA. \ett-s learns to utilize tokens from relevant temporal segments and tasks. The attention weights of AR tokens are large when the current action is indicative of future action. 
  }
  \label{fig:viz_lta}
\end{figure}

Our proposed \ett \ explicitly models task relations via a task translator and offers good interpretability on task relations. For \ett-s, Figure \ref{fig:viz_lta} shows the attention weights of task tokens when the primary task is LTA and the auxiliary task is AR. Given two adjacent input video clips, the goal of LTA is to predict the next action (\eg, put container and turn off nozzle for the two examples here). In the upper example, there is a scene change from the first clip (the temporal segment corresponding to put wheel) to the second clip (the clip corresponding to take container). The attention weights of AR tokens are small for the first clip and large for the second clip. Clearly, the future action to predict is more closely related to the second temporal segment due to similarities in the scene and objects. In the lower example, the AR tokens have large attention weights, as the video is temporally similar and the previous two actions are indicative of the next action. These results show how \ett-s accurately characterizes temporal and auxiliary task information to improve the primary task. More visualizations are in Appendix \ref{ap_sec:viz}. 

\begin{figure}[!t]
  \centering
  \includegraphics[width=1.0\linewidth]{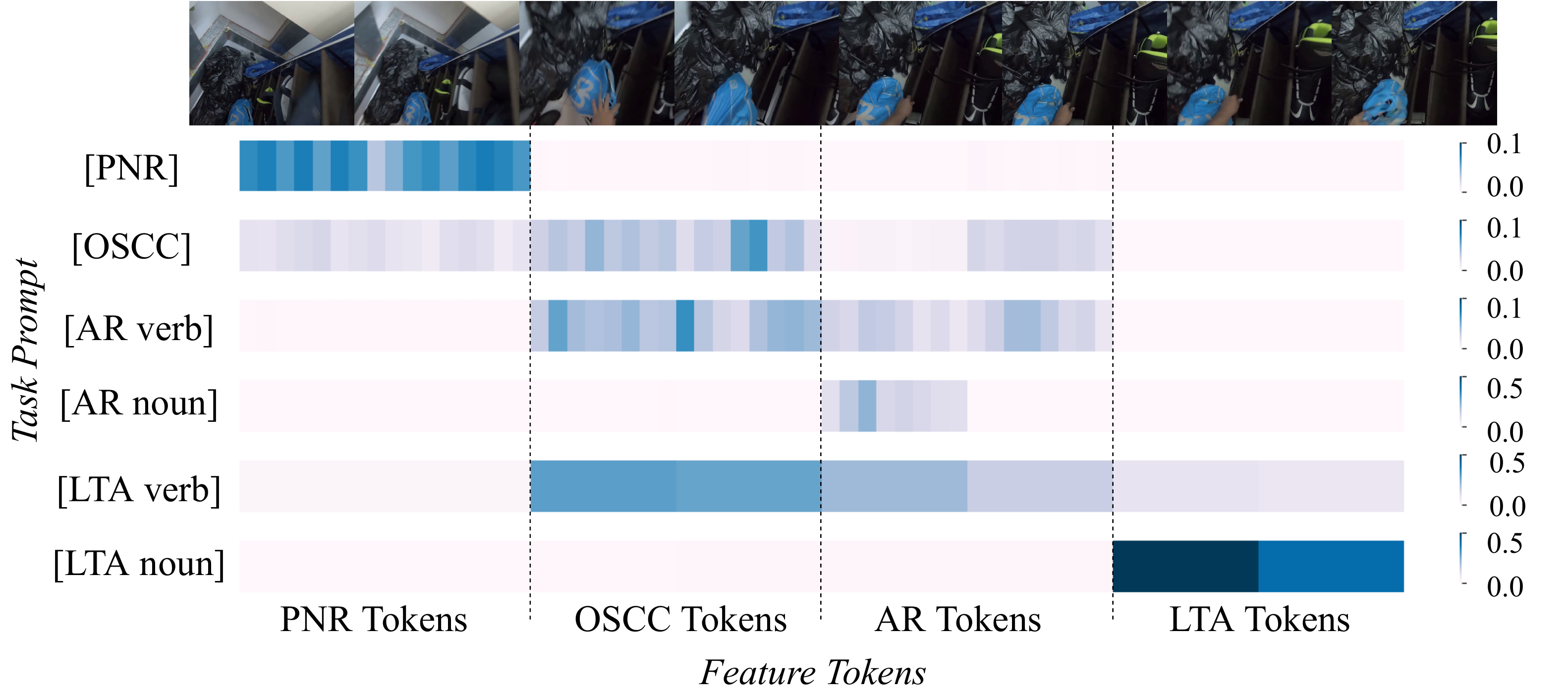}
  \caption{Attention weights of \ett-g. 
  Given the same video and different task prompts, \ett-g assigns different weights to different task tokens.  See text.}
  \label{fig:attn_weights_instance}
\end{figure}


Similarly, for \ett-g, we visualize its encoder-decoder attention weights from the last layer transformer in Figure \ref{fig:attn_weights_instance}. 
Given the same video clip as input, feature tokens are activated differently when \ett-g is given different task prompts, demonstrating that \ett-g learns to perform task translation conditioned on the task of interest. \revision{As it assigns small weights to task features that are not beneficial for the task of interest (\eg, PNR features to noun prediction tasks), \ett-g discards non-relevant task features to mitigate task competition.} 
We also observe temporal differences of attention weights from same task features, indicating that \ett-g captures both inter-frame and inter-task dependencies to improve the task of interest. 
\revision{Finally, recall that in Figure \ref{fig:intro}, we derive task relations for 4 human-object interaction tasks via attention weights provided by \ett-g. The attention weights are temporally pooled and averaged over all validation data, revealing task relations from a global perspective. Results for human-human interaction tasks are presented in Appendix \ref{ap_sec:viz}. In all, \ett \ provides good interpretability patterns on (1) which subset of tasks (2) which time segments lead to the final prediction. }


\section{Conclusion}
As a step towards unified egocentric perception, 
we propose \ett, a general and flexible design for task translation. \ett \ consists of heterogeneous video models optimized for each individual task and a transformer-based task translator that captures inter-frame and inter-task relations. 
We propose \ett-s to improve one primary task and \ett-g to conduct task translation for all tasks simultaneously.
Results on 7 diverse egocentric video tasks reveal valuable task relations and validate the proposed design.  





\newpage
{\small
\bibliographystyle{ieee_fullname}
\bibliography{egbib}
}

\newpage
\appendix

\ifarxiv
\section{Appendix}
This Appendix includes:
\begin{enumerate}
\item Reference to video with qualitative examples of \ett
\item Experimental Setup
\item Additional Results
\item Additional Visualizations
\end{enumerate}
\subsection{Video containing qualitative results}
We invite the reader to view the video available at \url{https://vision.cs.utexas.edu/projects/egot2/} where we show qualitative examples of (a) how \ett \ captures inter-frame and inter-task relations, (b) video retrieval results using attention weights of task tokens and (c) how \ett-g makes predictions conditioned on the task prompt and given video.
\fi

From these examples, we can see that \ett \ offers good interpretability on task relations, revealing clearly which temporal segments and which subsets of tasks contribute to improving a given task. Moreover, we run \ett-s on all AR validation videos and retrieve video segments with top PNR and OSCC weights. The results show that videos with large PNR and OSCC weights actually all involve heavy human-object interactions, which is the focus of these two tasks. Finally, we observe that \ett-g successfully performs task translation conditioned on the task of interest and task tokens through encoder-decoder attention weights. 

\subsection{Experimental Setup}
Below we provide detailed descriptions of the 7 tasks we adopt in our study.
\begin{itemize}
    \item \emph{Point-of-no-return Keyframe Localization (PNR)}: given a short video of a state change, estimate the keyframe that contains the time at which a state change begins.
    \item \emph{Object State Change Classification} (OSCC): given a video clip, classify whether an object state change has taken place or not.
    \item \emph{Action Recognition} (AR): classify the action (verb and noun) of the camera wearer from a short egocentric video clip; there are 115 verb categories and 478 noun categories.
    \item \emph{Long-term Action Anticipation} (LTA): given a video clip, predict the camera wearer's future sequence of actions; the action vocabulary is identical to that used in AR.
    \item \emph{Looking At Me} (LAM): given an egocentric video in which the faces of social partners have been localized and identified, classify whether each face is looking at the camera wearer.
    \item \emph{Talking To Me} (TTM): given a video and audio with the same tracked faces, classify whether each face is talking to the camera wearer.
    \item \emph{Active Speaker Detection} (ASD): given a cropped face video clip and corresponding audio segments, identify whether this person is speaking.
\end{itemize}

\subsubsection{Dataset Details}\label{ap_sec:data}
Note that Ego4D does not have a common training set that provides labels for all tasks. In all of our experiments, the task-specific models are trained on 7 subsets of Ego4D. Table~\ref{tab:dataset_percentage} reports the percentage of training videos shared between task pairs. Among the 7 task datasets, the average data overlap between two tasks is 22.5\%, and for 57.1\% of the task pairings there is strictly disjoint training data. The limited intersections in these 7 task datasets lend support to the generalizability of \ett \ across multiple datasets. 

\begin{table}[!h]
\centering
\scalebox{0.85}{
\begin{tabular}{cccccccc}
\toprule
& PNR & OSCC & AR & LTA & LAM & TTM & ASD \\
\midrule
PNR & 100 & 48.2 & 32.6 & 32.6 & 0 & 0 & 0\\
OSCC & 48.2 & 100 & 67.0 & 67.0 & 0 & 0 & 0\\
AR & 32.6 & 67.0 & 100 & 100 & 0 & 0 & 0 \\
LTA & 32.6 & 67.0 & 100 & 100 & 0 & 0 & 0\\
LAM & 0 & 0 & 0 & 0 & 100 & 12.8 & 12.8\\
TTM & 0 & 0 & 0 & 0 & 12.8 & 100 & 100\\
ASD & 0 & 0 & 0 & 0 & 12.8 & 100 & 100\\
\bottomrule
\end{tabular}}
\caption{Percentage of training videos shared between task pairs for the 7 tasks used in the experiments. There is a low level of data overlap between individual task pairs.}
\label{tab:dataset_percentage}
\end{table}

\subsubsection{Implementation Details}\label{ap_sec:implem}
\noindent{\bf{Task-Specific Translation}}. As shown in Table \ref{tab:dataset} of the main paper, the LTA task-specific backbone requires videos of 16 seconds while the other three human-object interaction tasks operate on videos of 8 seconds. Therefore, when $\mathcal{T}_p$ is LTA, we slide the other task-specific backbones along the 16-seconds time window to obtain auxiliary task features; the stride size is set to 8 seconds. When $\mathcal{T}_p$ is PNR, OSCC or AR, LTA is not a valid auxiliary task since its task-specific model requires video of a longer temporal span than provided in these three datasets. While it is possible to expand the video for the LTA model to be applicable, we aim at avoiding advantages brought by a longer time window for a fair comparison with prior work and thus exclude LTA as the auxiliary task. Nevertheless, to provide a complete evaluation, we consider one such special case when the primary task is AR and the auxiliary task is LTA (see results marked with $^\star$ in Table \ref{tab:ts_subset2}). Similarly, for the 3 human-human interaction tasks, LAM dataset provides video instances of 0.2 seconds while the TTM and ASD task-specific model requires videos of a longer time span. Consequently, LAM is not considered as the primary task. 

\begin{table*}[t]
  \centering
  \scalebox{0.85}{
  \begin{tabular}{lcc|cc|cc}
    \toprule
     & \multicolumn{2}{c}{$\mathcal T_p$ is TTM} & \multicolumn{2}{c}{$\mathcal T_p$ is PNR} & \multicolumn{2}{c}{$\mathcal T_p$ is OSCC} \\
     & \# Params $\cdot 10^6$ & mAP & \# Params $\cdot 10^6$ & Error  & \# Params $\cdot 10^6$ & Acc. \\
     & Trainable (\emph{All}) & (s) & Trainable (\emph{All}) & (s) $\downarrow$ & Trainable (\emph{All}) & (\%) $\uparrow$ \\
    \midrule
  TS model ~\cite{ego4d} & 20.2 \emph{(20.2)} & 58.91 & 32.2 \emph{(32.2)} & 0.615 & 32.2 \emph{(32.2)} & 68.22 \\
    \midrule
 \ett-s (Subset of Tasks) & 0.7 \emph{(35.3)} & 65.89 & 5.8 \emph{(70.2)} & \bf{0.608} & 5.8 \emph{(70.2)} & 69.69 \\
 \ett-s (All Tasks) & 0.7 \emph{(51.1)} & \bf{66.54} & 6.4 \emph{(132)} & 0.610 & 7.4 \emph{(133)} & \textbf{72.69} \\  
    \bottomrule
  \end{tabular}
 }
  \caption{Results of \ett-s when primary task is $\mathcal{T}_p$ is TTM, PNR and OSCC. We compare \ett-s that uses a subset of auxiliary tasks with \ett-s using all auxiliary tasks. When $\mathcal{T}_p$ is TTM, `Subset of Tasks' denote TTM and LAM; When $\mathcal{T}_p$ is PNR or OSCC, `Subset of Tasks' denote PNR and OSCC.}
  \label{tab:ts_subset1}
\end{table*}

\begin{table*}[t]
  \centering
  \scalebox{0.85}{
  \begin{tabular}{l|ccc|ccc}
    \toprule
     & \multicolumn{3}{c}{$\mathcal T_p$ is AR} & \multicolumn{3}{c}{$\mathcal T_p$ is LTA}   \\
     & \# Params $\cdot 10^6$ & \multicolumn{2}{c}{Acc. (\%) $\uparrow$} & \# Params $\cdot 10^6$ & \multicolumn{2}{c}{ED@20 $\downarrow$} \\
     & Trainable (\emph{All}) & Verb & Noun & Trainable (\emph{All}) & Verb & Noun \\
    \midrule
  TS model ~\cite{ego4d} & 63.3 \emph{(63.3)} & 22.18 & 21.55 & 180 \emph{(242)} & 0.746 & 0.789 \\
    \midrule
 \ett-s (Subset of Tasks) & 2.4 \emph{(282)} & 21.94$^\star$ & \bf{23.33}$^\star$ & 25.0 \emph{(304)}  & 0.739 & 0.774\\
     \ett-s (All Tasks) & 4.3 \emph{(130)} & \textbf{23.04} & 23.28 & 41.8 \emph{(348)} & \textbf{0.731} & \textbf{0.769} \\  
    \bottomrule
  \end{tabular}
 }
  \caption{Results of \ett-s when primary task is $\mathcal{T}_p$ is AR and LTA. `Subset of Tasks' denote AR and LTA. The results achieved with expanded video length are makred with a $^\star$.}
  \label{tab:ts_subset2}
\end{table*}

\noindent{\bf{Task-General Translation}}. For human-object interaction tasks, we follow common practices~\cite{huang2020mutual} and treat predicting verbs and nouns as two separate tasks. \ett-g is thus jointly optimized on 6 tasks: PNR, OSCC, AR Verb, AR Noun, LTA Verb and LTA Noun. We simplify the LTA task as predicting actions at a single timestamp into the future {as opposed to the 20 timestamps considered in the original benchmark} since otherwise the decoder would be heavily biased towards the LTA task (see parameter comparison in Table \ref{tab:ts_HOI} of the main paper). While \ett-s predicts future actions at future 20 timestamps and uses edit distance@20 (ED@20) as the metric, we report verb and noun accuracy for LTA in \ett-g.  
For human-human interactions, while LAM is not considered as the primary task for \ett-s, \ett-g provides the flexibility to incorporate LAM in training as well. In particular, when task prompt is LAM, we feed LAM tokens as input to the task fusion transformer and do not use other task tokens following the time span guidelines discussed above.  

\noindent{\bf{Tokenization and Detokenization}}. We construct a small task-related vocabulary for the sequence decoder in \ett-g to work. Namely, it is based on the label spaces of all candidate tasks and maps the original output label to a vocabulary. For PNR, we transform the output keyframe (\ie, an integer from 0-15) to be its character format. For OSCC/LAM/TTM/ASD, we transform the output label to the word `True' or `False'. For AR and LTA, we use the verb and noun vocabulary and transform the label to the word. In addition, we include the 7 task prompts (\ie, PNR, OSCC, AR, LTA, LAM, TTM and ASD) in the vocabulary. Consequently, we can transform the original label for all tasks to be a sequence and transform the predicted sentence back to the original label since it is a one-to-one mapping. Note that \ett-g is not sensitive to the choice of prompts. For example, the human-human interaction task prompts are [LAM], [TTM], [ASD], but [TaskA], [TaskB], [TaskC] would work too. Any output tokens outside the target task’s label space are considered incorrect predictions. We find that EgoT2-g learns to predict words within the target task dictionary after a few epochs.

\noindent{\bf{Hyperparameters and Optimization}}. Our implementation is based on the official Ego4D codebase.\footnote{\url{https://github.com/EGO4D}.} \ett-s retains the same training configurations (\eg, batch size, optimizer, total number of training epochs) unless otherwise specified. (1) $\mathcal{T}_p$ is PNR: Transfer (AR) is implemented as a SlowFast backbone pretrained on AR dataset followed by a 1-layer MLP with hidden dimension of 4096 and the PNR prediction head. Similarly, Finetuning and Transfer (OSCC) consists of a I3D ResNet-50 backbone pretrained on PNR and OSCC respectively followed by a 1-layer MLP with hidden dimension of 512 and the PNR prediction head. Late Fusion uses 3 1-layer MLPs to map features generated by each task-specific model (\ie, PNR, OSCC and AR) to be 512-dimensional and concatenates the three task-specific features; the concatenated features are then passed to the PNR prediction head. \ett-s consists of 6-layer transformer encoders with hidden dimension of 128. (2) $\mathcal{T}_p$ is OSCC: we follow the same way as in PNR to implement these baselines, and the task fusion transformer in \ett-s  has 5 layers with hidden dimension set as 128. (3) $\mathcal{T}_p$ is AR: Late Fusion follows the same design as in PNR and OSCC but has hidden dimension equal to 256. \ett-s uses a transformer encoder of 3 layers and hidden dimension set as 128. (4) $\mathcal{T}_p$ is LTA: The hidden dimension of Finetuning, Transfer and Late Fusion is set as 2048. \ett-s has a 1-layer transformer encoder with 128 dimension. (5) $\mathcal{T}_p$ is TTM: Finetuning and Transfer baselines are implemented as 3-layer MLPs with hidden dimension set as 1024 and 512. Late Fusion uses a 2-layer MLP to take concatenated features as input and passes the processed features to the TTM prediction head. \ett-s uses a 1-layer transformer encoder with hidden dimension of 128. (6) $\mathcal{T}_p$ is ASD: The baselines follow the same design as in TTM, and the hidden dimension of Transfer and Late Fusion is set as 6144 and 2048, respectively. \ett-s uses a 1-layer transformer encoder with hidden dimension of 256. Learning rate is set as 1e-3. 

For \ett-g on human-object interaction tasks, we use a batch size of $4 \times 8$ distributed over 8 GPUs. The task translator consists of 3 transformer encoder layers and 3 transformer decoder layers with hidden dimension equal to 512. We use AdamW optimizer with learning rate and weight decay set as 1e-4. For human-human interaction tasks, we set the batch size for LAM, TTM and ASD to be 256, 15 and 1800 respectively to balance three dataloaders. The task translator has 1 transformer encoder layer and 1 transformer decoder layer with hidden dimension set as 128. We use Adam optimizer with learning rate of 5e-4 and weight decay of 5e-5. All models are trained for 20 epochs.

\subsection{Additional Results}\label{ap_sec:result}

\noindent{\bf{Analysis on Task Relations}}. From Table \ref{tab:ts_HOI}-\ref{tab:ts_HHI} in the main paper, we observe the superior performance of \ett-s. Moreover, Transfer baseline results from these two tables offer insights on task relations. Intuitively, tasks within one benchmark (\eg, AR and LTA) are very related and can help each other, and tasks across benchmarks (\eg, PNR and AR, OSCC and AR) may seem unrelated at first sight. It is interesting to see that our results capture both inter-benchmark and intra-benchmark task relations: (1) when $\mathcal T_p$ is PNR, the Transfer of OSCC or AR features yields similar results, achieving the temporal localization error of 0.611 and 0.613 seconds, respectively; (2) when $\mathcal T_p$ is OSCC, surprisingly, Transfer (AR) outperforms Transfer (PNR) and a dedicated OSCC model (\ie, Finetuning) by $\sim$3\%; (3) when $\mathcal T_p$ is AR or LTA, PNR and OSCC features transfer better to predicting verbs than predicting nouns. We hypothesize that this is because an object state change is dependent on verbs and agnostic to nouns. 

We find that the task of action recognition (AR) is very informative in predicting the other 3 tasks; this suggests that similar to common practices in third-person video understanding (\eg, finetuning an action recognition model pretrained on Kinetics to other downstream tasks), the Ego4D AR model can also serve as a good initialization choice for other egocentric video tasks. In addition, from the task definition, PNR and OSCC are more object-centric while AR and LTA focus on human activities. Besides the obvious task relations (\ie, PNR to OSCC, AR to LTA), we uncover connections between tasks belonging to different benchmarks as well. AR task provides information complementary to primary task features and benefits OSCC. PNR and OSCC models convey information that are helpful for classifying verbs in AR and LTA. 

For human-human interactions, when the primary task is TTM, the good results achieved by Transfer (LAM) and Transfer (ASD) indicate that both auxiliary tasks provide informative cues for TTM. This also aligns with our intuition that LAM and TTM are very related tasks as people tend to make eye contact when they talk to someone. In addition, when $\mathcal T_p$ is ASD, Transfer baseline results indicate that TTM and LAM are detrimental to the ASD task. We conjecture that this may be because the act of someone looking at the camera wearer does not necessarily relate to the fact that this person is the active speaker. In all, we hope our analysis on task relations can facilitate holistic egocentric video understanding.

\noindent{\bf Varying the Set of Auxiliary Tasks.} In Table \ref{tab:ts_HOI}-\ref{tab:ts_HHI} of the main paper we presented results for \ett-s (All Tasks), where all tasks within the same cluster of $\mathcal T_p$ are adopted as auxiliary tasks. Here we consider the setting where we constrain the auxiliary tasks to be within the same benchmark as $\mathcal T_p$. Results of \ett-s using a subset of tasks\footnote{We exclude ASD here since there is no other task from the same benchmark as ASD (see Table \ref{tab:dataset} in the main paper).} are shown in Table \ref{tab:ts_subset1}-\ref{tab:ts_subset2}. 

By comparing results of \ett-s (Subset of Tasks) and \ett-s (All Tasks) in these two tables, we see that there are cases where \ett \ can effectively leverage synergies between tasks that belong to different benchmarks. For instance, when $\mathcal T_p$ is OSCC, since AR features provide beneficial cues, \ett-s with all auxiliary tasks outperforms  by 3\% the \ett-s variant that only uses PNR and OSCC features. Conversely, we would expect that the introduction of inter-benchmark auxiliary tasks may cause a detrimental effect when the benchmarks involve dissimilar tasks, for instance, when $\mathcal T_p$ is PNR. However, even in such case \ett-g (All Tasks) is still on-par with \ett-g (Subset of Tasks) and it outperforms all transfer baselines. This suggests that it has strong ability to mitigate negative transfer. 

\noindent{\bf Ablation Study.} In Table \ref{tab:ablation} of the main paper, we provided an ablation study of \ett-s when the primary task is TTM to validate our design choices. Here, we conduct another set of ablation studies for the case when $\mathcal{T}_p$ is OSCC. The results are summarized in Table \ref{tab:ablation_oscc}. The results are consistent with those reported in Table \ref{tab:ablation}. The three components (\ie, the introduction of auxiliary tasks, preserving temporal information and freezing TS backbones) work together and contribute to the efficacy of \ett-s. 

\begin{table}[t]
 \centering
 \scalebox{0.8}{
  \begin{tabular}{lccccc}
  \toprule
  & \makecell[c]{\# Params $\cdot 10^6$ \\ Trainable (\emph{All})} & \makecell[c]{Auxiliary\\ Tasks} & \makecell[c]{Temporal\\ Information} & \makecell[c]{Frozen\\ TS model} & \makecell[c]{mAP \\(\%) $\uparrow$} \\
  \midrule
  (a) & 8.9 \emph{(105)} & & $\checkmark$ & $\checkmark$ & 69.68\\
  (b) & 7.4 \emph{(133)} & $\checkmark$ &  & $\checkmark$ & 71.65\\
  (c) & 133 \emph{(133)} & $\checkmark$ & $\checkmark$ & & 72.22\\
  (d) & 7.4 \emph{(133)} & $\checkmark$ & $\checkmark$ & $\checkmark$ & 72.69\\
  \bottomrule
  \end{tabular}
}
\caption{Ablation study of \ett-s ($\mathcal{T}_p$ is OSCC). 
}
  \label{tab:ablation_oscc} 
\end{table}

\noindent{\bf Experiments with a different TS backbone.}
In the experiments presented in the main paper, we selected as TS backbones, the baseline models of Ego4D in order to facilitate comparison with prior work and to demonstrate the ability of our approach to achieve state-of-the-art results with simple network designs. However, \ett-s provides a flexible framework that can incorporate any advanced architecture. Here we demonstrate this flexibility by replacing the I3D ResNet-50 backbone with a video transformer used in EgoVLP~\cite{lin2022egocentric} for the case when $\mathcal{T}_p$ is OSCC. We report results in Table \ref{tab:result_egovlp}. We find that the improvement brought by auxiliary task information (\ie, AR in this case) is orthogonal to architecture advances and pretraining techniques. \ett-s can further improve the EgoVLP model performance by 2.77\%.

\begin{table}[t]
 \centering
 \scalebox{0.8}{
  \begin{tabular}{lccc}
  \toprule
  Acc. (\%) & SlowFast & EgoVLP \\
  \midrule 
 TS Model & 68.22 & 73.00\\
 \ett-s & 72.69 & 75.77\\
  \bottomrule
  \end{tabular}
}
\caption{Experiments with the TS model being SlowFast and EgoVLP when $\mathcal{T}_p$ is OSCC. By resorting to auxiliary task information, \ett-s demonstrates further performance improvements. 
}
  \label{tab:result_egovlp} 
\end{table}

\begin{figure}[t]
  \centering
  \includegraphics[width=0.85\linewidth]{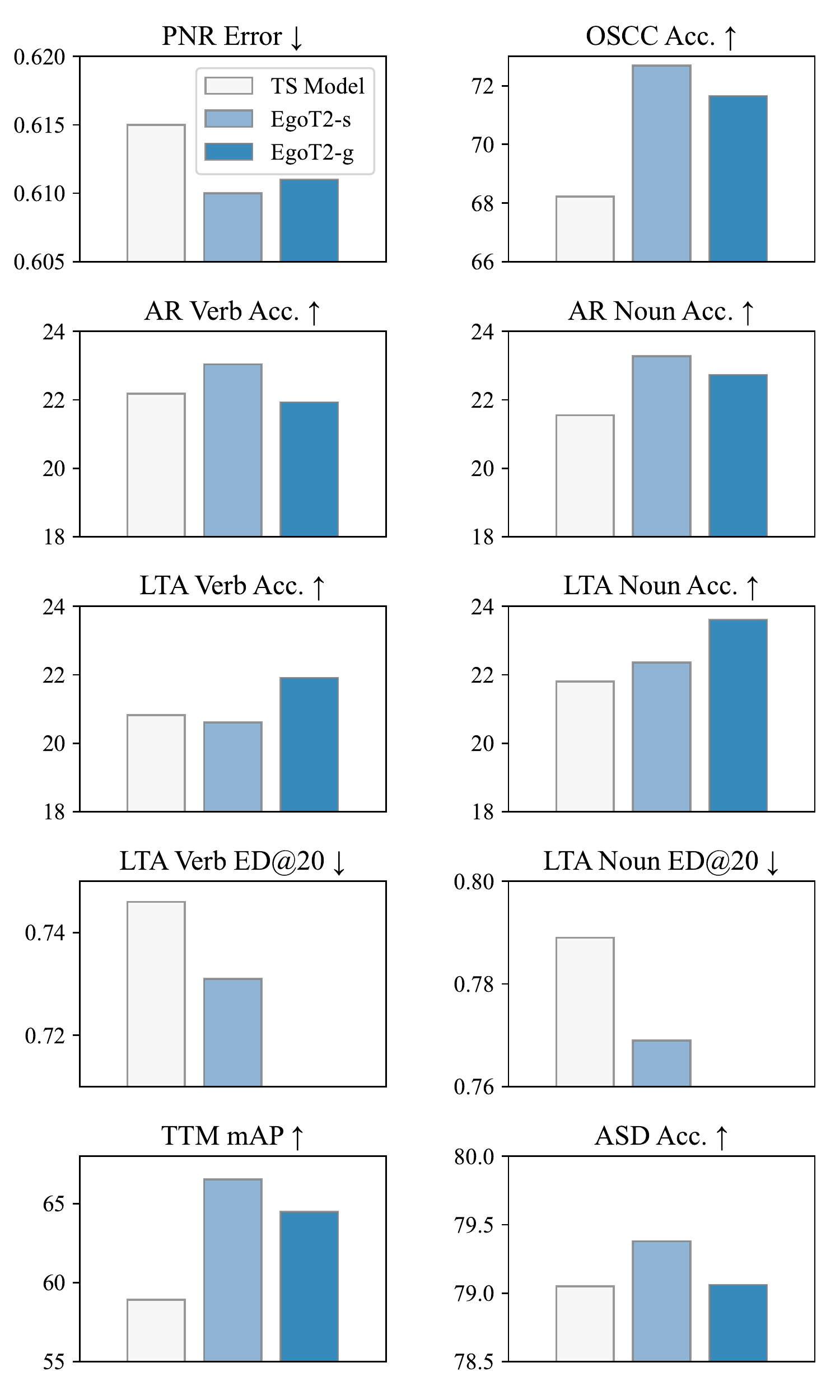}
  \caption{Performance comparison of two variants of \ett \ with the TS models on 6 tasks. \ett \ leads to great improvement over the TS model and \ett-s achieves top performance.}
  \label{fig:ett_cmp}
\end{figure}

\noindent{\bf Comparison of \ett-s and \ett-g.} We provide a side-by-side comparison of our proposed two variants of \ett \ over the TS model in Figure \ref{fig:ett_cmp}. As discussed in Sec. \ref{ap_sec:implem}, LTA has two metrics (accuracy for future 1 timestamp and edit distance for future 20 timestamps). Since \ett-s is optimized towards long-term predictions and \ett-g is trained to make one-step predictions, \ett-s does not perform as well as \ett-g in terms of LTA verb and noun accuracy, and ED@20 is not computable for \ett-g. In general, \ett \ achieves great performance gains over the TS models across tasks, and \ett-s leads to top performance. Moreover, Table \ref{tab:cmp_parameter} compares the number of trainable parameters and multiply-accumulate operations required for \ett-s and \ett-g. For \ett-s, we sum the trainable parameters (computations) of all task translators within one cluster. \ett-g shares the task translator across tasks within one cluster and hence saves parameters. The computational costs of \ett-s and \ett-g are similar, as the majority of the computation lies in the task-specific backbones, which are identical in both variants. 

\begin{table}[t]
 \centering
 \scalebox{0.8}{
  \begin{tabular}{lcccc}
  \toprule
  & \multicolumn{2}{c}{Human-Object Tasks} & \multicolumn{2}{c}{Human-Human Tasks}\\
   & \# Params & \# MACs &\# Params & \# MACs \\
\midrule
 Sum of \ett-s & 2.2 & 1802.5 & 59.9 & 386.6 \\
 \ett-g & 1.4 & 1803.6 & 34.5 & 386.2 \\
  \bottomrule
  \end{tabular}
}
\caption{Efficiency comparison of two variants of \ett. We report the number of trainable parameters (in millions) and the multiply-accumulate operations (MACs, in billions) required for one forward pass. Compared with a set of \ett-s models developed for each task, \ett-g has fewer trainable parameters and similar computational costs.}
  \label{tab:cmp_parameter} 
\end{table}

\noindent
\textbf{EgoT2-g across Task Clusters.} In Table \ref{tab:tg} of the main paper, we presented separate results of \ett-g on the cluster of human-human interaction (HHI) tasks and the cluster of human-object interaction (HOI) tasks due to the substantial domain gap between hese two clusters of Ego4D videos. Take Figure~\ref{fig:task} as an example, videos from HOI task datasets (upper figure) only capture human-object interactions and do not have other people in the scene. Thus, no HH interactions would be detected in these HOI videos, and as such we would not expect HHI features to contribute to HOI tasks. To verify this hypothesis, we implement a cross-cluster EgoT2-g that attends to two HOI tasks (PNR and OSCC) and one HHI task (LAM) simultaneously and report the results in Table \ref{tab:egot2g_cluster}. The cross-cluster \ett-g yields similar performance with intra-cluster \ett-g.

\begin{table}[h]
 \centering
 \scalebox{0.8}{
  \begin{tabular}{lccc}
  \toprule
    & \makecell[c]{PNR\\ Error (s) $\downarrow$} & \makecell[c]{OSCC \\Acc. (\%) $\uparrow$} & \makecell[c]{LAM\\ mAP (\%) $\uparrow$} \\
   \midrule
   \ett-g (intra-cluster) & 0.612 & 68.6 & 77.63 \\
   \ett-g (cross-cluster) & 0.611 & 68.3 & 77.56 \\
  \bottomrule
  \end{tabular}
}
\caption{Results of \ett-g across 2 task clusters. Due to the domain gap between human-human interaction tasks and human-object interaction tasks, \ett-g (cross-cluster) does not lead to further improvement compared with the \ett-g variant trained within the same task cluster.}
  \label{tab:egot2g_cluster} 
\end{table}


\subsection{Additional Visualizations}\label{ap_sec:viz}
\begin{figure}[t]
  \centering
  \includegraphics[width=1.0\linewidth]{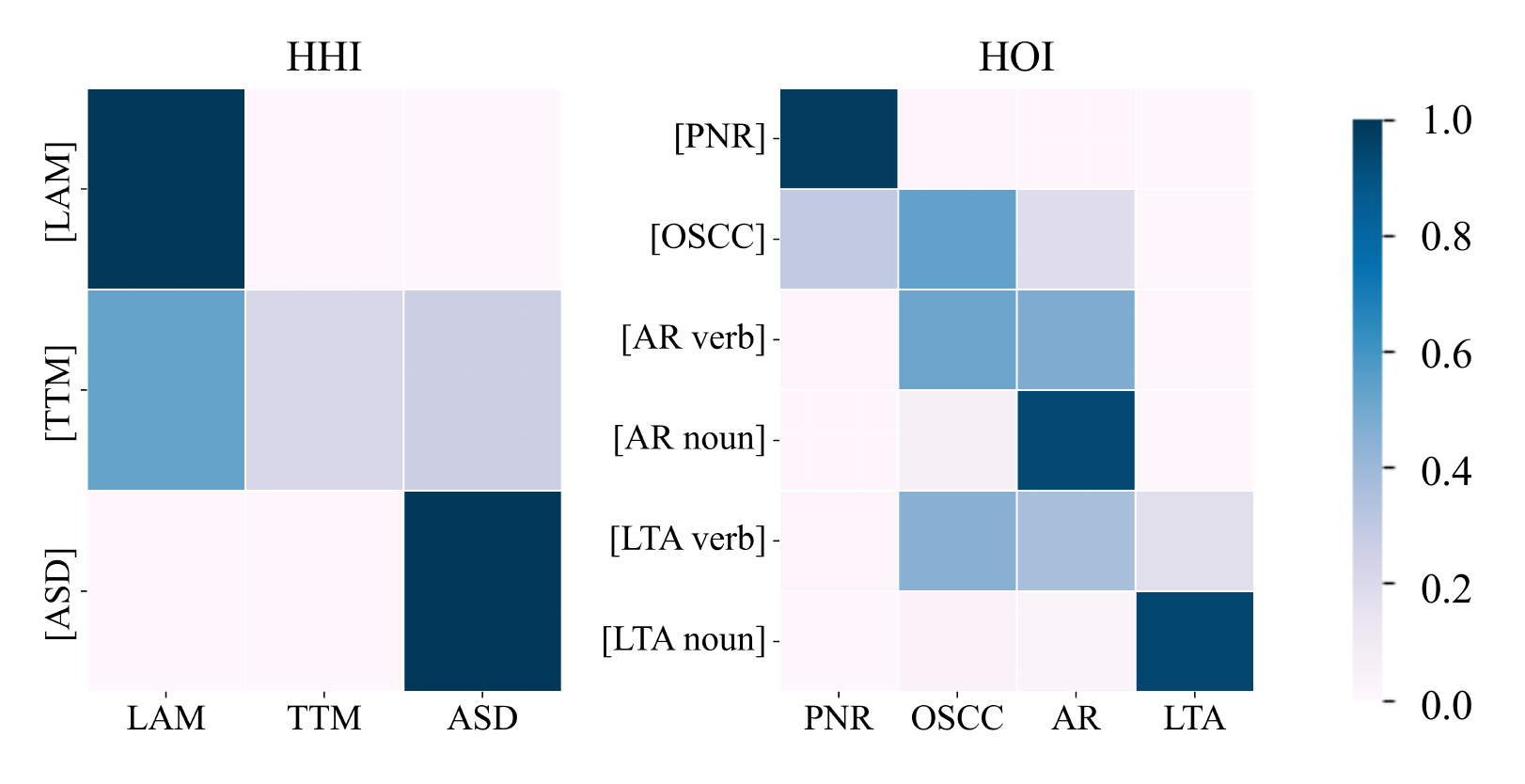}
  \caption{Average encoder-decoder attention weights of \ett-g. The heatmaps illustrate how task-specific feature tokens ($x$ axis) contribute to the task of interest ($y$ axis) in task translation.}
  \label{fig:attn_weights}
\end{figure}

Finally, Figure~\ref{fig:attn_weights} shows encoder-decoder attention weights of the last layer transformer produced by \ett-g for 3 human-human interaction (HHI) tasks and 6 human-object interaction (HOI) tasks. The attention weights of task-specific tokens are temporally pooled into one token and averaged over all validation video data. $x$ axis are different task tokens and $y$ axis corresponds to task prompts. Note that in Figure~\ref{fig:intro} in the main paper, we average the attention weights of verb and noun for AR and LTA and visualize the resulting $4\times4$ matrix. 
Figure~\ref{fig:attn_weights} reveals inherent task relations and provides an intuitive illustration of how the task-general translator utilizes task tokens differently conditioned on the task of interest (\ie, task prompt). In the left figure, we observe that LAM and ASD features have large attention weights when the task prompt is TTM, indicating that \ett-g effectively utilizes the two relevant tasks to improve TTM predictions. On the contrary, when the task prompt is ASD, ASD tokens are largely activated while non-beneficial LAM and ASD tokens are rarely adopted in task translation. This demonstrates that \ett-g learns to selectively activate task tokens to mitigate the issue of negative transfer. In the right figure, AR task tokens are more activated given that the task prompt is OSCC rather than PNR. This aligns with our previous finding in \ett-s that AR features are beneficial for the OSCC task. Also, when the task of interest is predicting nouns (\ie, task prompt is AR noun or LTA noun), attention weights of PNR and OSCC are very small, which indicates that the two task features do not help in noun prediction. The conclusion is also consistent with \ett-s.

\end{document}